# Disinformation Detection: A review of linguistic feature selection and classification models in news veracity assessments

Jillian Tompkins

State University of New York at Albany
College of Emergency Preparedness, Homeland Security and Cybersecurity
Department of Information Science
IIST 669
August 2018

# Table of Contents

## Introduction

Over the past couple of years, the topic of "fake news" and its influence over people's opinions has become a growing cause for concern. Although the spread of disinformation on the Internet is not a new phenomenon, the widespread use of social media has exacerbated its effects, providing more channels for dissemination and the potential to "go viral." Nowhere was this more evident than during the 2016 United States Presidential Election. Although the current of disinformation spread via trolls, bots, and hyperpartisan media outlets likely reinforced existing biases rather than sway undecided voters, the effects of this deluge of disinformation are by no means trivial. The consequences range in severity from an overall distrust in news media, to an ill-informed citizenry, and in extreme cases, provocation of violent action.[1] It is clear that human ability to discern lies from truth is flawed at best. As such, greater attention has been given towards applying machine learning approaches to detect deliberately deceptive news articles.

## Problem Statement

Automated deception detection is not a new area of research. Text classification problems related to detecting deceptive, misleading, or fraudulent information have been applied to use cases such as spam filtering, phishing alerts, online reviews and opinion spam, and fake social profiles. Only recently, however, have researchers begun to apply these classification problems to assess the veracity of news articles. The research problem to be solved is how effective these methods are in correctly classifying a news article as legitimate or fake.

---

[1] https://www.pbs.org/newshour/science/real-consequences-fake-news-stories-brain-cant-ignore

## Categories of Fake News

In order to address the fake news problem, it is necessary to first define fake news. While there is no universally agreed upon definition, the existing literature references several defining characteristics. (V. L. Rubin, Chen, & Conroy, 2015) define three categories of fake news, with their key differences resting on intent.

*Fabrications* are examples of fraudulent reporting. Yellow journalism, conspiracy theories, political propaganda and sensationalist clickbait articles fall into this category. The Hillary Clinton "Pizzagate scandal" is an example of a serious, widespread fabrication.

*Hoaxes* are another type of fabrication designed to deceive audiences and masquerade as real news. Hoaxes are particularly dangerous because they may get mistakenly picked up by traditional news outlets.

Unlike the previous two categories, *satire*, and other humorous fakes such as parody, are designed to serve as entertainment. These types of fakes do not carry the intention of deceiving its readers. Satirical content sites such as *The Onion* are clearly labeled as such. As we will see in a later section, satire shares several features with deceptive news articles.

## Theoretical Background

Part of what makes the news classification problem such a significant undertaking has to do with the sheer volume of information that is regularly produced and disseminated. It is practically infeasible for a human to manually classify and fact check every document. There needs to be automated processes in place to perform these classification tasks that are as reliable or, better yet, more reliable than human judgement. This is where machine learning comes into play.

Machine Learning is a subfield of Artificial Intelligence that applies algorithms that learn from examples instead of hardcoded procedural rules. This is important as it allows for greater reusability. Machine Learning can be broken down into two main types: unsupervised learning and supervised learning. Unsupervised learning involves the clustering of data and is most frequently used in order to detect patterns and anomalies. In unsupervised machine learning tasks, inferences are drawn from datasets that are unlabeled. In other words, there is only input (X) data and no corresponding output (Y) variables. This is in contrast to supervised learning tasks, where the training data consists of labeled input-output sets. The end goal of supervised learning is to take new, previously unseen input data and correctly assign its corresponding output variable.

Determining whether a news article is fake or legitimate is an example of a binary classification problem. Classification is one of two supervised machine learning problems, the other being regression. Classifiers take data as input and assign a label as output.

### Machine Learning Workflow

The workflow for a classification problem can be summarized as follows:

Gather annotated dataset → Perform necessary pre-processing (stemming/lemmatizing, stop word removal, normalization) → Feature selection and extraction → Split data into training and test sets (with optional validation sets) → Train classifier using training set → Test performance using test data set.

### Corpus Selection

(V. L. Rubin et al., 2015) have defined 9 corpora requirements best suited for text analysis of news articles. The most significant requirements can be summarized as follows:

*Balance of truthful and deceptive texts*. In order for any predictive model to be successful, it must be able to find patterns in both positive and negative data points.

*Homogeneity in article length*. The corpus should consist of documents of comparable lengths.

*Thematic homogeneity*. Genres and topics within a corpus should be aligned.

*Predefined timeframe*. Due to the evolving nature of language and news discourse, the corpus should be collected within a specific time frame.

Available Data Sets

The following publicly-available data sets are used in the studies referenced throughout this paper:

Buzzfeed Election Data Set[2]. This data set, gathered during the months leading up to the 2016 United States Presidential Election, is a collection of real and fake news stories with the highest Facebook engagement. Buzzfeed News gathered this data using keyword searches on the content analysis tool BuzzSumo (Horne & Adali, 2017).

Buzzfeed Hyperpartisan Facebook Page Dataset[3]. (Granik & Mesyura, 2017; Potthast, Kiesel, Reinartz, Bevendorff, & Stein, 2017). Not to be confused with the previous Buzzfeed dataset, this dataset contains a series of articles published on Facebook over the span of a week in late September 2016. Each article was fact-checked by 5 Buzzfeed journalists. The corpus includes 1,627 articles—828 from mainstream news agencies, 356 from left-wing sources, and 545 from right-wing sources.

---

[2] https://www.buzzfeednews.com/article/craigsilverman/viral-fake-election-news-outperformed-real-news-on-facebook
[3] https://www.buzzfeednews.com/article/craigsilverman/partisan-fb-pages-analysis; https://github.com/BuzzFeedNews/2016-10-facebook-fact-check

Political News Dataset (Horne & Adali, 2017). This dataset contains 75 stories from each of the three predefined news categories: fake, real, and satire. These stories were gathered from known sources for each category. Political stories were randomly selected from each of the sources. Opinion pieces were excluded from the dataset.

LIAR dataset (Wang, 2017). This dataset consists of 12,836 short statements collected from various contexts, including political debates, Facebook posts, tweets, and interviews. These statements were scraped from Politifact's API, with each statement evaluated by a Politifact editor for truthfulness. The labels used for this dataset are pants-fire, false, barely-true, half-true, mostly-true, and true.

ISOT Fake News Dataset[4] (Ahmed, Traore, & Saad, 2017). This dataset contains 44,848 articles that are almost evenly split between real (21,417) and fake (23,481). The real news articles all come from Reuters.com, while the fake articles come from various websites that were flagged as unreliable by Politifact and Wikipedia. The articles are thematically and temporally homogenous, focusing on political and world news from 2016 to 2017.

## Feature Selection

In order to train a classifier, it is necessary to select and extract a set of features that can be converted into numeric values. Since the classifiers are trained using data from the extracted features, proper selection of these features is a critical step. Redundant and irrelevant features can negatively impact the performance and accuracy of the classifier (Ahmed et al., 2017). The question then becomes, which features have the most predictive power? Are there inherent stylistic differences between legitimate and deceptive news articles? Although features used in classification problems range in scope, such as network-based and context-based, this paper focuses on linguistic feature selection only, and will be broken out into three subcategories: lexical, syntactic, and psycholinguistic.

---

[4] https://www.uvic.ca/engineering/ece/isot/datasets/

### Lexical Features

Lexical features are character-level and word-level features. Examples of lexical features include total word count, average word length, informal or slang words, and unique words. (Horne & Adali, 2017) compute word-level complexity using different grade-level readability indexes, which calculate a grade level score based on the number of syllables per word. The authors also use a metric called the Type Token Ratio (TTR) to capture the lexical diversity of a document.

### Syntactic Features

Syntactic features are sentence-level features. These include parts-of-speech (POS) tagging, parse tree depth, punctuation, and average sentence length. To capture sentence-level complexity, (Horne & Adali, 2017) use the Stanford Parser to compute each sentence's syntax tree depth. The deeper the tree depth, the greater the complexity of the sentence. Several authors (Conroy, Rubin, & Chen, 2015; S. Gilda, 2017; Shu, Sliva, Wang, Tang, & Liu, 2017) have experimented with deep syntax using probabilistic context free grammars (PCFGs) to detect deceptive language, with mixed results.

### Psycho-linguistic Features

A common characteristic of fake news articles is emotionally-charged, inflammatory language. Positive or negative sentiment, high occurrences of personal pronouns, and sensationalist (click-bait) language are all examples of psycho-linguistic features of fake news. (Horne & Adali, 2017) use a sentiment analysis tool called SentiStrength to measure the level of positive or negative sentiment intensity of each document. (Volkova & Jang, 2018) also extract psycholinguistic cues using the Linguistic Inquiry and Word Count (LIWC) program, such as imperative commands, personal pronouns, and emotional language.

### Choosing an Algorithm

Many different classification algorithms exist, each with their own strengths and weaknesses. To narrow the scope of this review, this paper will focus on two of the most common family of algorithms used in classification tasks: Naïve Bayes and Support Vector Machines.

### Naïve Bayes

Naïve Bayes is a popular probabilistic classification algorithm based on Bayes theorem of conditional probability. It is "naïve" in that it assumes that all features are independent of one another. In spite of its simplicity and perceived limitations, this algorithm tends to generalize well and is most commonly used in spam filtering. (Granik & Mesyura, 2017) chose to use a Naïve Bayes classifier based on the perceived similarities between fake news articles and spam emails. Among these common features include emotionally-charged language, grammatical errors, and lexical redundancy.

(Svärd & Rumman, 2017) compare the performance of two common Naïve Bayes algorithm implementations: Multinomial Naïve Bayes and Bernoulli Naïve Bayes. They tested each classifier on two test sets of different sizes—a smaller test of 10 real and 10 fake articles, and a larger test of 10 real and 100 fake articles.

### Support Vector Machines

Another popular algorithm used in supervised classification tasks are Support Vector Machines (SVMs). This classification algorithm is used in studies by (Ahmed et al., 2017; Horne & Adali, 2017; S. Gilda, 2017; Wang, 2017). Support Vector Machines construct a decision boundary known as a hyperplane in a high-dimensional space. The data points that fall along the margin of this hyperplane are referred to as the support vectors. In SVMs, the optimal hyperplane maximizes the margin between the classes. An advantage of this algorithm is that it performs well even when there is a high number of features (Ahmed et al., 2017). A disadvantage of this algorithm when compared to Naïve Bayes is that it is far more computationally intensive.

## Performance Evaluation

Evaluating a binary classifier's performance is a crucial step in measuring its predictive efficacy. Although several evaluation metrics exist, they are all related to measurements of precision and recall. These measurements can be calculated from the values in the confusion matrix, which is a two-dimensional table consisting of actual and predicted classes. Confusion matrices are commonly used to visualize the performance of a classifier and report the following values:

**True Positive (TP)** – predicted and actual class both positive (e.g. fake news classified as fake)

**True Negative (TN)** – predicted and actual class both negative (e.g. real news classified as real)

**False Negative (FN)** – incorrect prediction of negative class (e.g. fake news classified as real)

**False Positive (FP)** – incorrect prediction of positive class (e.g. real news classified as fake)

From these values we can measure the following:

**Precision** measures what proportion of predicted positives are true positives. Following the fake news example, precision would answer what proportion of predicted fake news articles were actually fake.

$$Precision = \frac{TP}{TP + FP}$$

**Recall** is a measurement that determines what proportion of actual positives were predicted as being positive. In the fake news example, this would translate to what proportion of fake articles did the classifier identify as fake.

$$Recall = \frac{TP}{TP + FN}$$

The **F1 score** is a measure that represents the harmonic mean of precision and recall.

$$F1\ score = 2 * \frac{Precision * Recall}{Precision + Recall}$$

**Accuracy** is the measure of correct predictions over all of the predictions.

$$Accuracy = \frac{TP + TN}{TP + TN + FP + FN}$$

Accuracy by itself is not a reliable measure of performance when there is class imbalance, which is often the case for news datasets where the majority of documents fall into one category over the other. A more useful metric to evaluate classifiers when there are unbalanced classes is to calculate the area under the ROC curve (ROC-AUC). The ROC curve plots the True Positive Rate (TPR), which is the same as the recall, against the False Positive Rate (FPR), which is:

$$FPR = \frac{FP}{FP + TN}$$

The ROC curve plots these two points at different classification thresholds. In order to get an aggregate measure across all classification thresholds, the area under the ROC curve is calculated. The ROC-AUC score ranges from 0 to 1, with 1 being a perfect classifier predicting correctly 100% of the time.

Using these metrics, the performance of classification algorithms can be evaluated.

### Naïve Bayes

(Svärd & Rumman, 2017) compared the performance of the multinomial Naïve Bayes and the Bernoulli Naïve Bayes classifiers on the two test sets. They found that the multinomial classifier outperformed the Bernoulli classifier on the small test set of 10 real and 10 fake articles, with an F1 score of 71.47%. The Bernoulli classifier outperformed the multinomial classifier on the larger test set of 10 real and 100 fake articles. The performance of both classifiers dropped significantly on the larger test set, showing the volatility of these classifiers.

(Granik & Mesyura, 2017) reported a classification accuracy score of 75.4%. Due to the skewedness of the dataset (with less than 5% of the articles labeled as fake), this evaluation should be taken with a grain of salt. The authors of this study acknowledge this fact as a limitation.

### Support Vector Machines

Horne and Adali's SVM classifier achieved a 71% accuracy when separating the text body of real and fake articles. The classifier yielded even greater accuracy when separating the titles of real and fake articles, suggesting that stylistic features of titles have strong predictive power. They found that fake titles are longer, more lexically redundant, contain fewer stop words, and use more capitalization. Fake titles also use more named entities and verb phrases, while real titles tend to be brief and general. The authors also used an SVM classifier to compare satire to fake and real news and found that satire has stylistically more in common with fake news than with real news. These similarities prove promising for deception detection research. (V. Rubin, Conroy, Chen, & Cornwell, 2016) used a SVM classifier that predicted satirical news with 90% precision and 84% recall. Although satire is not meant to be deliberately deceptive, it has the potential to deceive readers when taken out of context such as on news aggregators.

## Limitations and Considerations for Future Research

The biggest obstacle facing the fake news classification problem is the overall lack of labelled data sets. Assembling and annotating a corpus of news articles is a time consuming and manual task. Categorizing an article as truthful or deceptive requires careful analysis and domain expertise (Shu et al., 2017; Wang, 2017). Furthermore, there are seldom cases where an article contains nothing but entirely false claims. There are often grains of truth contained within mostly fabricated and misleading texts. As evidenced in the LIAR dataset, even short statements can contain varying degrees of truth.

Another limitation in several of the studies (specifically Granik & Mesyura, 2017; Svärd & Rumman, 2017) was the skewedness of the datasets. Larger datasets with a balanced representation of classes that also follow the corpus guidelines outlined by (V. L. Rubin et al., 2015) are needed.

There is currently not enough research to suggest that the use of linguistic features alone is sufficient for class prediction. More research needs to be done to evaluate linguistic feature used in tandem with other features, such as network and context-based features.

## Conclusion

Research of the application of machine learning techniques to news veracity assessments is still in the nascent stages, but the body of literature is growing. As more and more people turn to social media as a source for news, the greater the urgency to have reliable detection mechanisms in place to help stop the spread of disinformation. The browser extension B.S.

Detector[5] is one example of such a mechanism. One of the ethical tenets of the LIS community is to promote the responsible production and consumption of information. After all, a well-informed citizenry is the cornerstone of any functioning democracy. Information professionals can help tackle the fake news problem not only through the promotion of information literacy, but through continued research of automated deception detection.

## Bibliography

Ahmed, H., Traore, I., & Saad, S. (2017). Detection of Online Fake News Using N-Gram Analysis and Machine Learning Techniques (pp. 127–138). https://doi.org/10.1007/978-3-319-69155-8_9

Conroy, N. J., Rubin, V. L., & Chen, Y. (2015). Automatic deception detection: Methods for finding fake news. *Proceedings of the Association for Information Science and Technology*, *52*(1), 1–4. https://doi.org/10.1002/pra2.2015.145052010082

Granik, M., & Mesyura, V. (2017). Fake news detection using naive Bayes classifier. In *2017 IEEE First Ukraine Conference on Electrical and Computer Engineering (UKRCON)* (pp. 900–903). https://doi.org/10.1109/UKRCON.2017.8100379

Horne, B. D., & Adali, S. (2017). This Just In: Fake News Packs a Lot in Title, Uses Simpler, Repetitive Content in Text Body, More Similar to Satire than Real News. *ArXiv:1703.09398 [Cs]*. Retrieved from http://arxiv.org/abs/1703.09398

---

[5] http://bsdetector.tech/

Potthast, M., Kiesel, J., Reinartz, K., Bevendorff, J., & Stein, B. (2017). A Stylometric Inquiry into Hyperpartisan and Fake News. *CoRR*, *abs/1702.05638*. Retrieved from http://arxiv.org/abs/1702.05638

Rubin, V., Conroy, N., Chen, Y., & Cornwell, S. (2016). Fake News or Truth? Using Satirical Cues to Detect Potentially Misleading News. https://doi.org/10.18653/v1/W16-0802

Rubin, V. L., Chen, Y., & Conroy, N. J. (2015). Deception Detection for News: Three Types of Fakes. In *Proceedings of the 78th ASIS&T Annual Meeting: Information Science with Impact: Research in and for the Community* (pp. 83:1–83:4). Silver Springs, MD, USA: American Society for Information Science. Retrieved from http://dl.acm.org/citation.cfm?id=2857070.2857153

S. Gilda. (2017). Evaluating machine learning algorithms for fake news detection. In *2017 IEEE 15th Student Conference on Research and Development (SCOReD)* (pp. 110–115). https://doi.org/10.1109/SCORED.2017.8305411

Shu, K., Sliva, A., Wang, S., Tang, J., & Liu, H. (2017). Fake News Detection on Social Media: A Data Mining Perspective. *ArXiv:1708.01967 [Cs]*. Retrieved from http://arxiv.org/abs/1708.01967

Svärd, M., & Rumman, P. (2017). *COMBATING DISINFORMATION : Detecting fake news with linguistic models and classification algorithms*. Retrieved from http://urn.kb.se/resolve?urn=urn:nbn:se:kth:diva-209755

Volkova, S., & Jang, J. Y. (2018). Misleading or Falsification: Inferring Deceptive Strategies and Types in Online News and Social Media. In *Companion Proceedings of the The Web Conference 2018* (pp. 575–583). Republic and Canton of Geneva, Switzerland:

International World Wide Web Conferences Steering Committee.

https://doi.org/10.1145/3184558.3188728

Wang, W. Y. (2017). "Liar, Liar Pants on Fire": A New Benchmark Dataset for Fake News Detection. *ArXiv:1705.00648 [Cs]*. Retrieved from http://arxiv.org/abs/1705.00648